# Risk-Aware Vehicle Trajectory Prediction Under Safety-Critical Scenarios

Qingfan Wang, Dongyang Xu, Gaoyuan Kuang, Chen Lv, Shengbo Eben Li, and Bingbing Nie

*Abstract*—Trajectory prediction is significant for intelligent vehicles to achieve high-level autonomous driving, and a lot of relevant research achievements have been made recently. Despite the rapid development, most existing studies solely focused on normal safe scenarios while largely neglecting safety-critical scenarios, particularly those involving imminent collisions. This oversight may result in autonomous vehicles lacking the essential predictive ability in such situations, posing a significant threat to safety. To tackle these, this paper proposes a risk-aware trajectory prediction framework tailored to safety-critical scenarios. Leveraging distinctive hazardous features, we develop three core risk-aware components. First, we introduce a risk-incorporated scene encoder, which augments conventional encoders with quantitative risk information to achieve risk-aware encoding of hazardous scene contexts. Next, we incorporate endpoint-risk-combined intention queries as prediction priors in the decoder to ensure that the predicted multimodal trajectories cover both various spatial intentions and risk levels. Lastly, an auxiliary risk prediction task is implemented for the ultimate risk-aware prediction. Furthermore, to support model training and performance evaluation, we introduce a safety-critical trajectory prediction dataset and tailored evaluation metrics. We conduct comprehensive evaluations and compare our model with several SOTA models. Results demonstrate the superior performance of our model, with a significant improvement in most metrics. This prediction advancement enables autonomous vehicles to execute correct collision avoidance maneuvers under safety-critical scenarios, eventually enhancing road traffic safety.

Manuscript received XX XX, XXXX; revised XX XX, XXXX; accepted XX XX, XXXX. Date of publication XX XX, XXXX; date of current version XX XX, XXXX. This work was supported by National Natural Science Foundation of China under Grant 52131201 and 52072216. Computing resource is supported by Center of High-Performance Computing, Tsinghua University. (*Corresponding authors: Bingbing Nie.*)

Qingfan Wang is with the State Key Laboratory of Intelligent Green Vehicle and Mobility, School of Vehicle and Mobility, Tsinghua University, Beijing 100084, China, also with the School of Mechanical and Aerospace Engineering, Nanyang Technological University, 639798, Singapore. (e-mail: wqf20@mails.tsinghua.edu.cn).

Dongyang Xu, Gaoyuan Kuang, Shengbo Eben Li, and Bingbing Nie are with the State Key Laboratory of Intelligent Green Vehicle and Mobility, School of Vehicle and Mobility, Tsinghua University, Beijing 100084, China (e-mail: xdy22@mails.tsinghua.edu.cn; kgy22@mails.tsinghua.edu.cn; lishbo@tsinghua.edu.cn; nbb@tsinghua.edu.cn).

Chen Lv is with the School of Mechanical and Aerospace Engineering, Nanyang Technological University, 639798, Singapore. (e-mails: lyuchen@ntu.edu.sg).

Color versions of one or more figures in this article are available at https://doi.org/ XXX.

Digital Object Identifier XXX

## I. INTRODUCTION

In intricate road traffic scenarios, accurately forecasting the multimodal trajectories of surrounding agents presents a formidable challenge for autonomous vehicles. As a research hotspot in the field of intelligent traffic [1], [2], precise and timely trajectory prediction is paramount for developing the subsequent decision-making system, including planning and control, eventually promoting the universal application of autonomous driving technology.

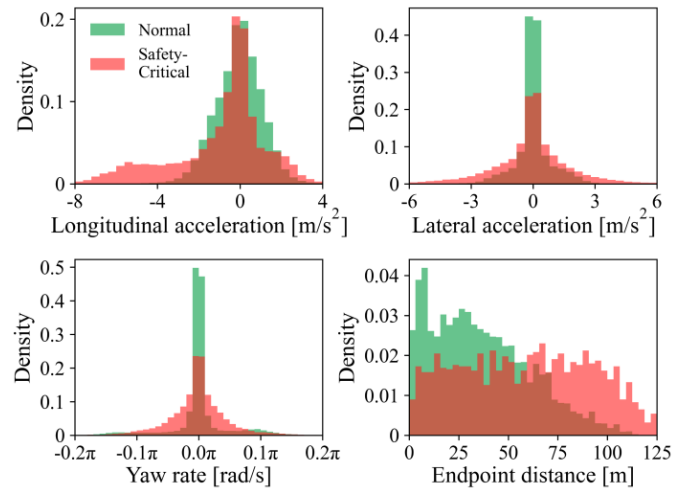

**Fig. 1.** Comparison of vehicle dynamics between normal scenarios (from WOMD [3]) and safety-critical scenarios (see Section V. A for more details).

Studies on trajectory prediction have progressed rapidly and proposed various prediction structures, resulting in the best prediction scores being constantly updated across different benchmarks [4], [5], [6], [7]. However, existing endeavors primarily concentrate on normal safe traffic scenarios, neglecting safety-critical scenarios, especially those involving imminent collisions. Meanwhile, we observe significant data heterogeneity between normal and hazardous driving behaviors, which can be reflected in the distribution discrepancies of vehicle dynamics, shown in Fig. 1. Compared with normal scenarios, vehicles' motion under safety-critical scenarios presents a more dispersed distribution, where behaviors involving large acceleration and yaw rate were more frequently observed. This means when confronting sudden traffic hazards, drivers tend to adopt more aggressive driving behaviors to avoid collisions. Recent studies have proposed various model structures to explore the interaction relationship between agents and maps since road lanes play an essential role in restricting vehicle behaviors [8], [9], [10], which have been used to facilitate accurate trajectory prediction. However, under safety-critical scenarios, such restrictions become less strict as collision avoidance becomes drivers' most important goal rather than obeying the traffic rules and arriving at the destination. Due to such nonnegligible heterogeneity in both vehicle dynamics and drivers' intentions, even state-of-the-art models (e.g., MTR, Motion transformer, ranking first place at Waymo Motion Prediction leaderboard [10]) fail to exhibit satisfactory trajectory prediction performance under intricate safety-critical scenarios. Quantitative evaluations, as depicted in Table I, underscore the noticeable performance degradation.

TABLE I
PREDICTION PERFORMANCE OF MTR UNDER NORMAL SCENARIOS AND SAFETY-CRITICAL SCENARIOS

|  | minADE (k=6[a]) | minFDE (k=6) | MR (k=6) |
|---|---|---|---|
| Normal Scenarios[b] | 0.516 | 1.040 | 0.123 |
| Safety-Critical Scenarios[c] | 1.940 | 5.186 | 0.357 |

a. k=6 means the minimal error among six predicted trajectories.
b. The prediction performance of MTR was tested on the validation set of the Waymo Open Motion Dataset [3].
c. MTR was re-trained and tested on our proposed safety-critical trajectory prediction dataset, of which details can be found in Section V. A.

We learn that existing widely used public trajectory prediction datasets, such as the Waymo open motion dataset (WOMD) [3], [11] and the Argoverse dataset [12], [13], were manually curated to exclude scenarios with potential hazards. Although such safety-critical scenarios belong to low-probability events, these rare cases still constitute a unique modality in real-world traffic. To achieve higher prediction accuracy, models trained on such datasets tend to consider surrounding agents to be 'friendlier' by predicting their trajectories with a lower likelihood of inducing accidents. In particular, some recent studies call it "social-aware" or "social-compatible" prediction [9], [14], [15], [16], [17]. We argue that though this makes sense in the majority of traffic scenarios and enhances prediction performance, it still poses inherent risks of traffic accidents in real-world deployment, potentially jeopardizing occupant safety by disregarding the dangerous trajectory modalities of surrounding vehicles.

To address the above issues, this study proposes a risk-aware trajectory prediction framework that can assess driving risks promptly and generate trajectory predictions aligned with prevailing risk levels. Such advancement is achieved by three core components tailored to safety-critical characteristics: a risk-incorporated scene encoder, a trajectory decoder guided by endpoint-risk-combined intention queries, and auxiliary risk prediction. Moreover, to mitigate the dearth of traffic hazards in existing public datasets, we develop an algorithm to automatically generate virtual safety-critical scenarios in CARLA and then conduct extensive driving simulation experiments to establish a safety-critical trajectory prediction dataset. Additionally, realizing the inadequacy of existing metrics in evaluating multimodal trajectory predictions under safety-critical scenarios, a series of specialized metrics are proposed to comprehensively assess the effect of predicted trajectories on subsequent risk judgment and safety decision-making of autonomous vehicles.

The contributions of this study are summarized as follows:
1) A risk-aware trajectory prediction framework is developed, which incorporates various core components to adapt to the characteristics of safety-critical scenarios.
2) We propose a trajectory prediction dataset that collects drivers' collision-avoidance responses under safety-critical scenarios by conducting large-scale driving simulation experiments. Meanwhile, a set of evaluation metrics tailored to safety-critical trajectory prediction tasks are developed.
3) We experimentally validate our risk-aware trajectory prediction framework on the safety-critical dataset. Compared with several SOTA models, our approach achieves the best trajectory prediction performance on both classical metrics and tailored safety-critical metrics.

This paper is structured as follows: First, recent studies related to models, datasets, and metrics of trajectory prediction are summarized in Section II. Then, Section III elaborates on the problem statement and the formulation of the proposed risk-aware trajectory prediction framework. Section IV details the establishment of the driver-centered risk indicator. Section V introduces our proposed training dataset and evaluation metrics designed for safety-critical scenarios. Experiment Setup and result analyses are conducted in Section VI, which are then discussed and concluded in Section VII.

II. RELATED WORK

As a connecting module between perception and planning, trajectory prediction has long been considered crucial to the ability of autonomous vehicles to understand traffic scenes, thus attracting significant research interest over the past years [18], [19], [20]. To achieve accurate trajectory prediction, researchers have carried out research from different aspects, including the prediction method itself, the training dataset, and the evaluation metrics. Here, a brief review summarizes the existing research from the above three aspects.

*A. Models for Trajectory Prediction*

Recently, various trajectory prediction frameworks have been proposed, constantly updating the best performance across different public benchmarks. These models typically adopt a hierarchical encoder-decoder architecture [21], [22], [23], [24], [25]. Given the manifold traffic factors influencing the vehicle's future motion, such as road geometry and surrounding agents' historical behaviors, recent studies elaborated different structures of the encoder to process these diverse factors and aggregated information via specific feature fusion techniques to achieve accurate scene understanding, e.g., the early fusion strategy for input modalities with factorized attention proposed by Wayformer [4] and the query-centric scene encoding paradigm proposed by QCNet [5]. Leveraging high-dimensional representations processed by the encoder, the decoder forecasts multiple future trajectories of the target agent along with their probabilities, i.e., modeling the multimodal motion patterns. For instance, MTR used a set of learnable motion query pairs as spatial intention priors to facilitate multimodal prediction [10].

However, most studies only focused on normal and safe scenarios. Trajectory prediction tasks under safety-critical scenarios remain unaddressed. Given the vital importance of predicted trajectories on subsequent decision-making of autonomous vehicles [26], a specialized trajectory prediction framework tailored to safety-critical scenarios is urgently needed, which can be developed by incorporating risk-aware components to quantify and infer the evolution of traffic risks and make corresponding predictions. Benefiting from such risk-aware trajectory predictions, autonomous vehicles can better identify risks and make reasonable collision-avoidance behaviors to mitigate potential collision risks.

*B. Dataset for Trajectory Prediction*

It is widely acknowledged that data quality plays a pivotal role in determining the performance of data-driven algorithms. Numerous trajectory prediction datasets have emerged to facilitate training algorithms [27], [28], [29]. Early datasets were typically small in scale and exhibited limited trajectory accuracy. Representative datasets include NGSIM [30] and HighD [31], which were collected from a specific segment of highways using roadside cameras or stationary drones. Currently, prominent trajectory prediction datasets, including Argoverse [12], [13], WOMD [3], [11], and Nuscene [32], have significantly enhanced the scale of the trajectory data. In addition to trajectory, they also offer detailed map information. However, during data collection, these datasets exclusively provide normal traffic data under safe scenarios, deliberately excluding safety-critical scenarios, such as collision and near-collision cases. This discrepancy from real-world traffic data can result in subpar performance of SOTA trajectory prediction models when tested under safety-critical scenarios. In application, such prediction errors could lead to misjudgments of the risk level of current scenarios by subsequent decision-making algorithms, finally inducing traffic accidents. The Interaction dataset [33] and Waymo interaction dataset paid more attention to the complex interactions between different traffic participants, alleviating the above issues to some extent. Nevertheless, they still lack traffic hazards that require traffic agents to react in a very short amount of time, which can be seen as a kind of 'extremely interactive' behavior. Datasets in traffic anomaly detection, e.g., DoTA [34] and PSAD [35], involve large-scale safety-critical scenarios but generally can only provide first-person videos or raw EDR data. Due to the lack of precise trajectory data, such datasets cannot support the training of trajectory prediction models. Overall, currently, a dataset consisting of large-scale trajectory data under safety-critical scenarios is urgently needed to support the training of risk-aware trajectory prediction models.

*C. Metrics for Trajectory Prediction*

The field of trajectory prediction has proposed several metrics to evaluate the deviation between actual and predicted trajectories [36]. Among these, the most prominent are average displacement error (ADE) and final displacement error (FDE), which quantify the average distance and endpoint distance between predicted and actual trajectories in Euclidean space. Given the diversity in drivers' intentions, recent studies have developed multimodal prediction by outputting multiple trajectories (and their probabilities) per prediction. This derives metrics, such as minADE and minFDE, representing the error of the best trajectory among the multiple predicted trajectories, formulated as:

$$\text{minADE} = \min_i \frac{1}{T_f} \sum_{t=1}^{T_f} \|\tau_{i,t}^p - \tau_t^{gt}\|_2$$
$$\text{minFDE} = \min_i \|\tau_{i,T_f}^p - \tau_{T_f}^{gt}\|_2 \tag{1}$$

where $\tau_{i,t}^p$ and $\tau_t^{gt}$ denote the $i$-th predicted trajectory and the ground-truth trajectory at the timestamp of $t$, $T_f$ is the future prediction frames, and $\|\cdot\|_2$ represents the L2 norm. Metrics like mean average precision (mAP) and brier-minFDE additionally consider different prediction confidences to enhance the assessment comprehensiveness. These existing metrics measure prediction errors in terms of trajectory itself. However, under safety-critical scenarios characterized by extremely short agent-agent distances, the same trajectory prediction errors could yield diametrically opposed outcomes, such as collisions and non-collisions. This highlights the imperative to account for the impact of predicted trajectories on driving risk and collision probability to achieve a more comprehensive evaluation under safety-critical scenarios.

III. RISK-AWARE TRAJECTORY PREDICTION

This section presents the detailed formulation of the proposed risk-aware trajectory prediction framework tailored to safety-critical scenarios (Fig. 2).

*A. Problem Statement*

The core goal of trajectory prediction is to accurately predict future positions (generally in 2D coordinates) of the target vehicle $\tau_{i,0:T_f}^p$, given its historical trajectory and observed surroundings $S = \{A, M\}$. $S$ includes other agents' historical trajectories $A_{-T_h:0}^{0:N_a}$ and map information $M_{0:P}^{0:N_m}$, represented by lane segments, where $T_f$ represents the future prediction frames, $T_h$ is historical frames, $i$ denotes the $i$-th mode of multimodal trajectory prediction, $N_a$ is the number of other observed agents, $N_m$ denotes the number of surrounding lane segments, and $P$ is the number of points to represent a lane segment's centerline.

For trajectory prediction under safety-critical scenarios, we emphasize the importance of quantifying risk to achieve risk-aware prediction. Thus, the surroundings $S$ are expanded as $S = \{A, M, R\}$, where the collision risk information between the target vehicle and other agents $R_{-T_h:0}^{0:N_a}$ is estimated based on an artificial potential field method, detailed in Section IV.

*B. Basic Architecture*

The design of our model was inspired by a previous SOTA trajectory prediction model, i.e., MTR [10], [37], which adopted a classical encoder-decoder architecture. Different from the original MTR, we incorporated a series of innovative

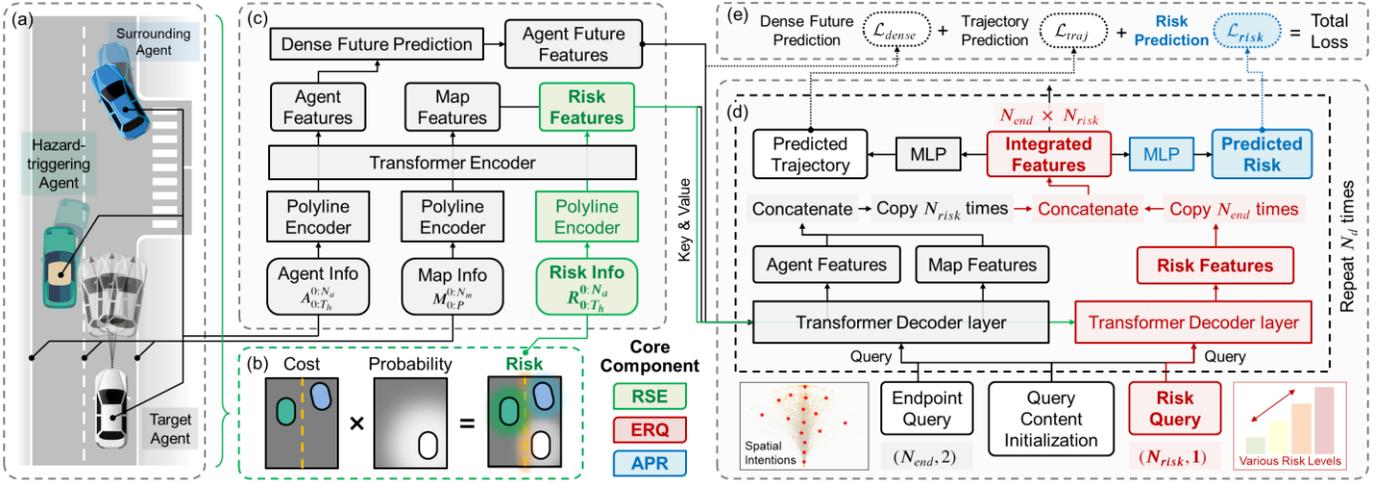

Fig. 2. The overall architecture of the proposed risk-aware trajectory prediction framework. Three core risk-aware components tailored to safety-critical scenarios are marked in color: risk-incorporated scene encoder (RSE), endpoint-risk-combined intention query (ERQ), and auxiliary prediction of risk (APR). (a) A representative safety-critical scenario, where the unreasonable lane change by the green vehicle poses a traffic hazard. (b) A brief diagram of the proposed artificial potential field-based risk quantification method. (c) A scene encoder that encodes the information of agent, map, and risk simultaneously. (d) A trajectory decoder that predicts future trajectories from the encoded scene information. (e) The whole framework is trained in an end-to-end fashion with three losses: trajectory prediction loss $\mathcal{L}_{traj}$, risk prediction loss $\mathcal{L}_{risk}$, and dense future prediction loss $\mathcal{L}_{dense}$

risk-aware components to cater to the safety-critical trajectory prediction task. Before detailing them, here we first briefly introduced the basic architecture of our model, which was similar to the original MTR.

**Scene Encoder.** The scene encoder first utilized polyline encoders, developed based on PointNet [38], to encode the raw inputs of agents $A_{0:T_h}^{0:N_a}$ and maps $M_{0:P}^{0:N_m}$ into high-dimension representations, which were then concatenated together. Next, a standard Transformer encoder was applied to further mine the coupling relationship between different inputs to achieve a more comprehensive encoding of scene context. To capture the potential future interactions between different agents, a multilayer perceptron (MLP)-based dense future prediction module was used to densely predict the trajectories of all agents, which was taken as an auxiliary supervised learning task with the loss denoted as $\mathcal{L}_{dense}$.

**Trajectory Decoder.** Taking the encoded agent and map features as key and value, a Transformer decoder consisting of $N_d$ cross-attention layers were applied to propagate information among different features. To generate multimodal predictions, the original decoder in MTR used 64 motion pairs, i.e., trajectory endpoints clustered by a k-means clustering algorithm, as the spatial intention queries. At last, MLP-based prediction heads predicted the future motions, including trajectory (each trajectory point is represented as a distribution modeled by a Gaussian mixture model), velocity, and their probabilities, which were used to calculate the final trajectory loss $\mathcal{L}_{traj}$, defined as:

$$\mathcal{L}_{traj} = \mathcal{L}_{reg} + \mathcal{L}_{cls} = \text{GR}(\tau_j^p, \tau^{gt}) + \text{CE}(p_{1:N}^p, j)$$
$$j = \arg\min_i \text{FDE}(\kappa_i, \tau^{gt}) \tag{2}$$

where $\text{GR}(\cdot,\cdot)$ denotes the Gaussian regression loss, $\text{CE}(\cdot,\cdot)$ is the cross-entropy loss, $\tau_j^p$ and $\tau^{gt}$ denote the $j$-th predicted trajectory and the ground-truth trajectory, respectively, $p_{1:N}^p$ is the predicted probability of all $N$ trajectory predictions, $\kappa_i$ is the $i$-th endpoint intention, and $j$ represents the sequence number of the endpoint intention with minimal errors from the ground-truth endpoint.

### C. Core Risk-Aware Components

Though MTR performed well on several public trajectory prediction datasets, its prediction performance under safety-critical scenarios still needs to be further enhanced. Therefore, we incorporated various architecture modification and training techniques into the original MTR to help it better interpret safety-critical scenarios, identify the potential risk level, and predict risk-aware multimodal trajectories. Technically, the core components of the proposed risk-aware model include: a risk-incorporated scene encoder, endpoint-risk-combined intention queries, and auxiliary risk prediction.

**Risk-incorporated Scene Encoder (RSE).** As described above, we introduced the quantitative traffic risk information $R_{0:T_h}^{0:N_a}$ as an independent input in addition to the conventional agent and map information. After being encoded by three polyline encoders $\text{PolyEnc}(\cdot)$, agent features $F_A$, map features $F_M$, and risk features $F_R$ were concatenated and passed together to a standard Transformer encoder with $N_{enc}$ stacked multi-head self-attention layers, where the risk features could further interact with the agent and map information to help the model enhance the understanding of safety-critical scene context (Fig. 2c). The whole process can be represented as:

$$F_{A/M/R} = \text{PolyEnc}\left(A_{0:T_h}^{0:N_a}/M_{0:P}^{0:N_m}/R_{0:T_h}^{0:N_a}\right)$$
$$E_{i+1} = \text{SelfAttn}(q = E_i + \text{PE}(E_i), k = E_i + \text{PE}(E_i), v = E_i), \quad i = 0,1,\dots,N_{enc}-1 \tag{3}$$
$$E_A, E_M, E_R = \text{Div}(E_{N_{enc}})$$

where $E_0 = [F_A, F_M, F_R] \in \mathbb{R}^{(N_a+N_m+N_a)\times D_f}$ represents the concatenated input features that aggregate all the scene context features, $D_f$ is the feature dimension, $[\cdot,\cdot]$ is the concatenation operation, $q$, $k$, and $v$ denote the query, key, and value in the self-attention module, and PE($\cdot$) denotes the sinusoidal position encoding of input tokens. The final outputs of the encoder $E_{N_{enc}}$ was divided (Div($\cdot$)) into the agent, map, and risk features $E_A \in \mathbb{R}^{N_a \times D_f}$, $E_M \in \mathbb{R}^{N_m \times D_f}$, and $E_R \in \mathbb{R}^{N_a \times D_f}$, which would be taken as the independent key and value for the subsequent independent cross-attention modules in the trajectory decoder to provide scene context information.

**Trajectory decoder with Endpoint-Risk-combined intention Queries (ERQ).** Our proposed trajectory decoder contains $N_{dec}$ decoder layers, one of which is depicted in Fig. 3. Each layer adopts two parallel Transformer decoder layers (gray box) to independently process trajectory-related features (blue box) and risk-related features (red box), which are then integrated to make final predictions.

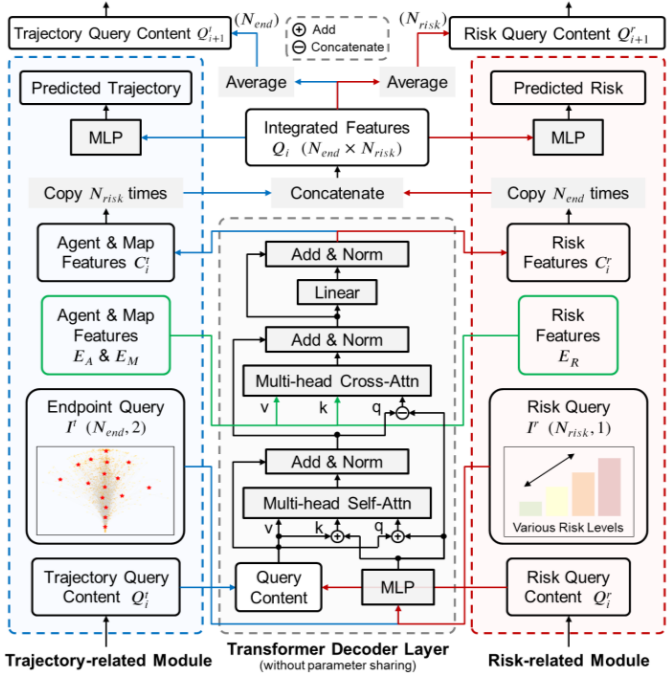

**Fig. 3.** The architecture of the $i$-th layer of the proposed trajectory decoder with endpoint-risk-combined intention queries. Blue and red boxes denote the modules designed for the tasks of trajectory prediction and risk prediction, respectively. The middle gray box represents a standard Transformer decoder layer shared by both tasks (without parameter sharing). Green boxes represent the encoded scene context features ($E_A$, $E_M$, and $E_R$) from the encoder.

Inspired by taking the clustered trajectory endpoints as spatial intention priors in MTR, we innovatively proposed to incorporate risk intentions to represent the potential risk level of the future scenario as priors. We expected this modification could instruct our model to make more risk-aware trajectory predictions. Technically, we first utilized MLPs to encode both endpoint intention queries $I^t$ (i.e., $N_{end}$ two-dimensional coordinates) and risk intention queries $I^r$ (i.e., $N_{risk}$ scalars) into high-dimension representations $T^{t/r}$:

$$T^{t/r} = \text{MLP } \text{PE}(I^{t/r}) \quad (4)$$

Then, two multi-head self-attention modules were used to process trajectory-related and risk-related features separately, where $T^t$ and $T^r$ were taken as the position embedding for queries and keys, developed as:

$$S_i^{t/r} = \text{SelfAttn}(q = Q_i^{t/r} + T^{t/r}, k = Q_i^{t/r} + T^{t/r}, v = Q_i^{t/r}), \quad i = 1,2,\ldots,N_{dec}-1 \quad (5)$$

where $Q_i^{t/r} \in \mathbb{R}^{N_{end}/N_{risk} \times D_f}$ denotes the trajectory-related and risk-related query content features at the $i$-th layer of the decoder, respectively. The query content features at the zeroth layer $Q_0^{t/r}$ are initialized as zeros.

For trajectory-related modules, the encoded information from self-attention $S_i^t$ and endpoint intentions $T^t$ were taken as the query in two cross-attention modules to interact with the agent features $E_A$ and map features $E_M$ from the encoder, respectively, formulated as:

$$C_i^{A/M} = \text{CrossAttn}(q = [S_i^t, T^t], k = [E_{A/M}, \text{PE}(E_{A/M})], v = E_{A/M}) \quad (6)$$

$$C_i^t = \text{MLP}([C_i^A, C_i^M])$$

where the outputs of cross-attention $C_i^A$ and $C_i^M$ were then aggregated into $C_i^t$ by an MLP. On the other hand, the risk-related modules also adopted a cross-attention module to query the encoded risk features $E_R$:

$$C_i^r = \text{CrossAttn}(q = [S_i^r, T^r], k = [E_R, \text{PE}(E_R)], v = E_R) \quad (7)$$

Note that the mode number of trajectory-related features $C_i^t \in \mathbb{R}^{N_{end} \times D_f}$ and risk-related features $C_i^r \in \mathbb{R}^{N_{risk} \times D_f}$ is different. To integrate them for the final predictions, we first extended their modes by copying trajectory-related features $N_{risk}$ times to get $C_i^{t,e} \in \mathbb{R}^{(N_{end} \times N_{risk}) \times D_f}$ and copying risk-related features $N_{end}$ times to get $C_i^{r,e} \in \mathbb{R}^{(N_{risk} \times N_{end}) \times D_f}$. Then, we utilized an MLP to aggregate them into an integrated feature, denoted as $Q_i \in \mathbb{R}^{(N_{end} \times N_{risk}) \times D_f}$. The above process can be described as:

$$C_i^{t,e} = \text{Ext}(C_i^t, N_{risk}), \quad C_i^{r,e} = \text{Ext}(C_i^r, N_{end})$$
$$Q_i = \text{MLP}([C_i^{t,e}, C_i^{r,e}]) \quad (8)$$
$$\tau^{p,i} = \text{MLP}(Q_i), \quad \gamma^{p,i} = \text{MLP}(Q_i)$$

where Ext($\cdot,\cdot$) denotes the extension by copying and $\tau^{p,i}$, $\gamma^{p,i}$ represent the predicted trajectories and risks at the $i$-th layer, both with the prediction modality of $N_{end} \times N_{risk}$. By incorporating endpoint-risk-combined intention queries, we established a two-dimensional prediction matrix, effectively ensuring that the multimodal trajectories predicted by our model not only covered various spatial intention priors (represented by trajectory endpoints) but also took into consideration different risk levels of the future traffic scene.

For the next $(i+1)$-th decoder layer, the trajectory-related and risk-related query content features $Q_{i+1}^{t/r}$ were obtained by averaging the corresponding dimensions of the integrated feature $Q_i$ to narrow down the number of predicted modes and restore them to their original dimensions, formulated as:

$$Q_{i+1}^t = \text{Avg}(Q_i, d = N_{end}) \\ Q_{i+1}^r = \text{Avg}(Q_i, d = N_{risk}) \quad (9)$$

where $Q_{i+1}^{t/r} \in \mathbb{R}^{N_{end}/N_{risk} \times D_f}$, and $\text{Avg}(\cdot, d)$ denotes the average operation with the dimensions $d$ to retain.

To iteratively update the predictions, the above process was repeated $N_{dec}$ times in the trajectory decoder. Via such iterative optimization, trajectory-related features and risk-related features were effectively integrated, not only facilitating the information interaction between endpoint intentions and risk intentions but also promoting the information propagation within each type of intention. A detailed analysis of the effect of the proposed risk intention queries on trajectory prediction under safety-critical scenarios is given in Section VI. E.

**Auxiliary Prediction of Risk (APR).** In addition to the conventional trajectory prediction task, we advocated simultaneously predicting the future evolution of traffic risks as an auxiliary supervised learning task to help the model identify the current risk levels. Compared with conventional models, the proposed risk-incorporated scene encoder and risk intention queries endowed our risk-aware model with the ability to accurately predict future risks. Upon obtaining the integrated feature $Q_i$ with $N_{risk} \times N_{end}$ modes, an independent MLP was adopted to predict the future risk sequence of the target vehicle $\gamma^p$, which is formulated in (8).

To promote multimodal predictions during optimization, the auxiliary risk prediction loss $\mathcal{L}_{risk}$ was calculated in a similar way as the trajectory prediction loss $\mathcal{L}_{traj}$, i.e., a hard-assignment strategy that calculates losses based on the risk intention query with minimal errors, formulated as:

$$\mathcal{L}_{risk} = \left\| \gamma_j^p - \max(\gamma^{gt}) \right\|_1 + \text{CE}(p_{1:N}^p, j) \\ j = \arg\min_i \left\| \nu_i - \max(\gamma^{gt}) \right\|_1 \quad (10)$$

where $\|\cdot\|_1$ denotes the L1 norm, $\gamma_j^p$ and $\gamma^{gt}$ are the $j$-th predicted risk sequence (in three dimensions, i.e., probability, cost, and risk) of the target agent and its ground-truth risk (calculated by the risk quantification method in Section IV), $p_{1:N}^p$ is the predicted probability of $N$ risk predictions, $\nu_i$ represents the $i$-th risk intention, and $j$ represents the sequence number of the risk intention with minimal errors from the maximum value of the ground-truth risk sequence.

Note that, for brevity, we simplified the formulation of $\mathcal{L}_{traj}$ and $\mathcal{L}_{risk}$ in (2) and (10). In fact, for loss calculation, the selection of predicted trajectories and risks from multimodal predictions is coupled. Technically, we first selected the optimal endpoint intention with the index of $i$ and the optimal risk intention with the index of $j$, independently. Then, the $(i, j)$-th prediction (including both trajectory and risk) in the coupled prediction matrix $(N_{risk}, N_{end})$ was selected as the optimal prediction to calculate losses in (2) and (10).

To sum up, the overall training loss $\mathcal{L}$ consists of three terms (Fig. 2e): trajectory prediction loss $\mathcal{L}_{traj}$, dense motion prediction loss $\mathcal{L}_{dense}$, and risk prediction loss $\mathcal{L}_{risk}$.

## IV. DRIVER-CENTERED RISK QUANTIFICATION

Numerous indicators have been proposed to quantify traffic risks, the most classic of which is time-to-collision (TTC) and time headway (THW). Since our ultimate goal is to predict vehicle trajectories determined by the driver's decision-making behaviors, an indicator that can assess the level of traffic risk from the driver's perspective is needed to assist in trajectory prediction. Thus, inspired by a previous study [39], we developed an artificial potential field-based traffic risk indicator. Technically, traffic risk perceived by drivers can be represented as the product of two parts: the probability of a collision and its cost (Fig. 2b) [40].

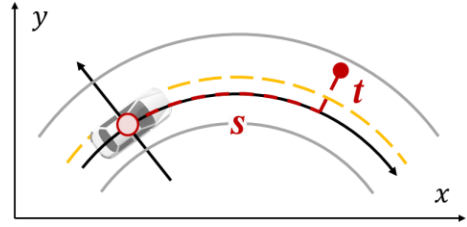

(a) A brief diagram of the coordinate transformation based on the arcing curve (under the assumption of constant velocity). $s$: longitudinal position. $t$: lateral position.

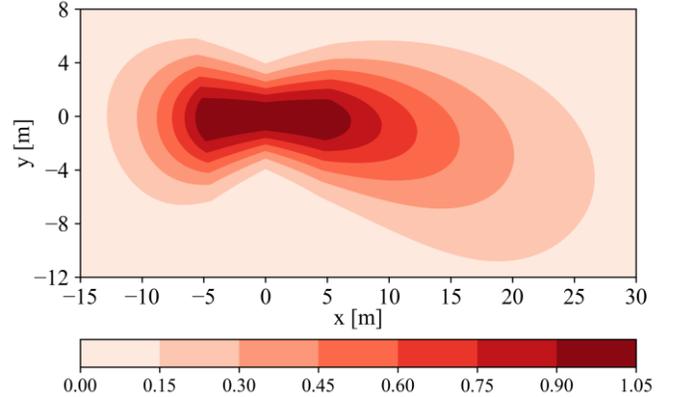

(b) An example of DRF (normalized to 0 to 1) with a velocity of 10m/s and a turning radius of 50m.

**Fig. 4.** Illustration of the driver's risk field.

### A. Collision Probability

The driver's subjective belief about the probability of a crash accident occurring was quantified via a method named driver's risk field (DRF) [39]. The DRF is represented as a two-dimensional matrix, where the target agent is located in the middle, and the value of each other cell stands for its collision probability with the target if an obstacle exists (Fig. 4). To model it, we assume that the target vehicle maintains a constant velocity and angular velocity in the coming few seconds. Thus, based on the target vehicle's current states, e.g., positions, velocities, and steering angle, we can utilize a planar 3-DoF bicycle model to estimate its future motion, which is represented as an arcing curve. Then, the position coordinates of each point in the DRF are projected onto this arcing curve. The new coordinate contains two parameters, i.e., longitudinal position $s$ and lateral position $t$. Following the principle that the closer the obstacle is, the greater the risk, the DRF was modeled as a torus with a Gaussian cross-section with varying height ($a$) and width ($\sigma$), which can be formulated as:

$$\text{DRF} = a(s) \cdot \exp\left(\frac{-t^2}{2\sigma^2(s)}\right)$$

$$a(s) = \begin{cases} A \cdot \dfrac{(s_{\max} - \max(s, s_{\min}))^2}{(s_{\max} - s_{\min})^2}, & s < s_{\max} \\ 0, & s \geq s_{\max} \end{cases} \quad (11)$$

$$\sigma^2(s) = (B \cdot \max(s, s_{\min}) + C)^2$$

where $s_{\max}$ is the maximum longitudinal position affected by the risk receptive field of drivers and is assumed velocity dependent, i.e., $s_{\max} = D \cdot (\max(v, 5))^E$, $s_{\min}$ is the allowed minimal value of $s$, and $A\sim E$ are constant coefficients.

### B. Collision Cost

The cost of a potential collision refers to the collision severity and occupant injury risks, affected by various factors, including vehicle dynamics, usage of restraint systems, and occupant characteristics. For simplification, we quantified the cost based on the impact energy at the vehicle level, which mainly contains three parts: a constant basic cost $E_b$ for collisions with any obstacle, an absolute cost equal to the obstacle's kinetic energy $E_a$ depending on its mass and absolute velocity, and a relative cost $E_r$ depending on the obstacle's relative velocity towards the target vehicle. The total collision cost equals the weighted sum of the three.

### C. Comprehensive Risk

After multiplying the above two parts, a comprehensive driver-centered risk indicator was obtained. Note that the risks posed to the target vehicle by different agents were calculated independently. For the target vehicle, the overall risk equals to the sum of all agents' risks, i.e., $R^0 = \sum_{i=1}^{N_a} R^i$. After concatenation, the final risk information $R^{0:N_a}$ with three dimensions, i.e., probability, cost, and comprehensive risk, was obtained. By integrating the agent and map information in traffic scenarios from the driver's perspective, the risk information $R$ is regarded as a vital auxiliary input for the trajectory prediction model to enhance its ability of scene understanding and risk perception, eventually assisting the model in making risk-aware trajectory predictions.

## V. DATASET AND METRICS TAILORED TO SAFETY-CRITICAL SCENARIOS

In addition to the risk-aware trajectory prediction model, training data and evaluation metrics tailored to safety-critical scenarios are also critical to improving prediction performance, which are detailed in this section.

### A. Dataset

For the establishment of the safety-critical trajectory prediction dataset, given the danger and low frequency of safety-critical scenarios in real-world traffic, we conducted a large-scale driving simulation experiment to ensure the safety and efficiency of data acquisition (Fig. 5).

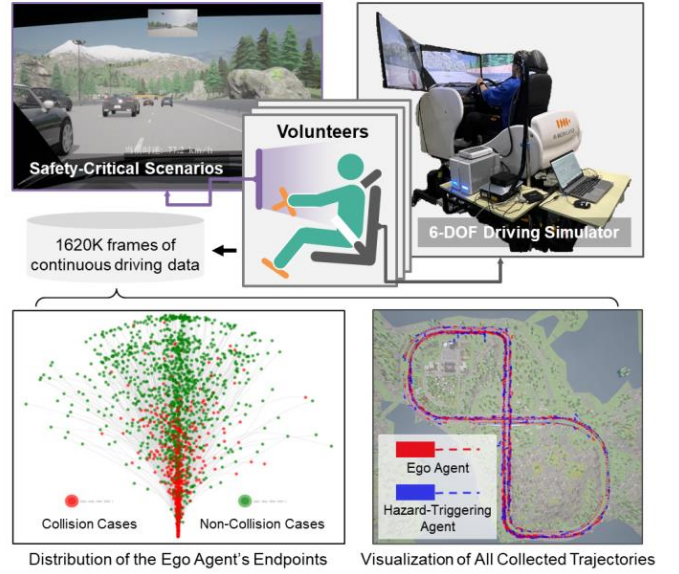

**Fig. 5.** Establishment of the safety-critical trajectory prediction dataset.

In terms of hardware, we utilized a high-fidelity driving simulator with a fully instrumented vehicle cab mounted on a 6-DOF motion system with three monitors to provide immersive traffic stimulation. Regarding software, various safety-critical events were designed based on the open-source simulation platform CARLA [41]. According to NHTSA [42], our experiment concentrated on three of the most frequent vehicle-to-vehicle conflict types, i.e., rear-end, cut-in, and merging scenarios (Fig. 6). We developed a hazard-triggering algorithm to achieve the random automatic generation of safety-critical events. Technically, this algorithm first assessed the collision risk between the ego vehicle controlled by the volunteer and surrounding background vehicles (controlled by our algorithm). The most aggressive background vehicle with the highest risk was selected as the hazard-triggering vehicle to execute aggressive behaviors and create traffic conflicts. Volunteers' various collision-avoidance behaviors subject to the random traffic hazards were automatically collected to form the final safety-critical trajectory prediction dataset after a series of necessary post-processing procedures. Each hazard-triggering instance was followed by a minimum 30-second interval, allowing drivers to revert to natural driving states.

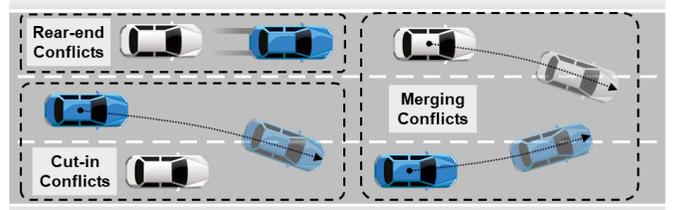

**Fig. 6.** Three types of designed traffic conflicts.

Twelve volunteers, aged between 22 and 43 years, with an average annual driving distance of more than 8,000 km, participated in the experiment, except one volunteer withdrew due to motion sickness. Volunteers underwent four 30-minute driving sessions, with 10-minute rest between sessions. Participants were free to select their routes and were responsible for executing evasive maneuvers when necessary. The Institutional Review Board (IRB) of Tsinghua University has approved the whole experimental procedure.

Via the experiment, we established a dataset consisting of 1620K frames of continuous driving data (22.5 hours with 20Hz) with detailed trajectory and map information. Among this data, 2088 safety-critical events were extracted, each covering five seconds before and after the hazard occurrence, which is defined as the moment when the hazard-triggering vehicle shows an obvious abnormal behavior, such as emergency braking at a deceleration exceeding a certain threshold and a reckless lane change with the lane offset exceeding a certain threshold. To increase the diversity of prediction cases, threshold values under different safety-critical scenarios vary, following Gaussian distribution. After removing anomalies, over 350K frames of 1879 safety-critical events are available for training and validating the proposed prediction framework. Table II summarizes the distributions of collected events' conflict type, collision time, and driver's emergency maneuvers when confronting safety-critical events. For more details about the dataset, please refer to another paper our team is currently preparing.

TABLE II
BASIC STATISTICAL INFORMATION ABOUT THE PROPOSED SAFETY-CRITICAL TRAJECTORY PREDICTION DATASET.

| Conflict Type | Number | Collison Time | Number | Maneuver Type | Number |
|---|---|---|---|---|---|
| Cut-in Conflicts | 1132 (60%) | Non-Collision | 1162 (62%) | Braking then Steering | 770 (41%) |
| Merging Conflicts | 331 (18%) | Collision in 1s | 264 (14%) | Steering then Braking | 376 (20%) |
| Rear-end Conflicts | 416 (22%) | Collision in 2s | 252 (13%) | Steering Only | 633 (34%) |
| - | - | Collision in 5s | 201 (11%) | Braking Only | 100 (5%) |

*B. Metrics*

Most existing trajectory prediction metrics measure errors in terms of the trajectory itself, i.e., distance errors between predicted and ground-truth trajectories. Under safety-critical scenarios, the predicted trajectories of surrounding agents will directly determine the ego autonomous vehicle's subsequent decision-making, eventually affecting the collision risk and the potential occupant injury severity. Thus, we advocate further considering the effects of trajectory prediction on evaluating the collision probability and severity.

Technically, for collision samples in the safety-critical dataset, prediction errors of relative collision velocity between the two vehicles were calculated to represent the estimation accuracy of collision severity. Prediction errors of the collision time were also calculated. Additionally, for safety-critical trajectory prediction, we emphasize that it is hazardous to predict a collision as a non-collision accident, which might cause autonomous vehicles to misjudge the collision risk and fail to make timely and reasonable avoidance behaviors. Thus, for collision samples, we calculated the collision miss rate ($MR_{coll}^k$). We defined that a collision miss occurs when all the multimodal predicted trajectories do not coincide with the surrounding vehicles' trajectories. Regarding non-collision samples, we evaluated the risk prediction errors based on the driving risk indicator proposed in Section IV to represent collision probability.

Similar to trajectory-related metrics, we also adopted mean square error and miss rate to calculate errors in predicting collision velocity (i.e., $MSE_{velo}^k$ and $MR_{velo}^k$), collision time (i.e., $MSE_{time}^k$ and $MR_{time}^k$), and traffic risk (i.e., $MSE_{risk}^k$ and $MR_{risk}^k$), where $k = 6$ denotes the multimodal prediction with six trajectories, and $k = 1$ denotes the unimodal prediction by selecting the predicted trajectory with the maximum predicted probability. $MR_{velo}^k$, $MR_{time}^k$, and $MR_{risk}^k$ were defined as the ratio of the validation scenarios where the prediction error of collision velocity, collision time, and traffic risk exceeds a threshold of 2.5 m/s, 0.25 s, and 0.15 (after normalization), respectively.

In addition, for existing trajectory-related metrics (e.g., minADE and min FDE), we split the samples in the validation dataset into four groups depending on the time of the collision (non-collision, collision-in-1s, collision-in-3s, and collision-in-5s) and calculated the evaluation results independently since the prediction time horizon significantly influences these metrics' magnitude.

VI. EXPERIMENTS

This section presented and discussed the experimental setup and evaluation results. We first introduced the validation experiments and implementation details. Second, the results of our proposed risk-aware trajectory prediction model are compared with those of several baseline methods. Finally, necessary ablation studies were conducted to demonstrate how each module in our model contributes to the final result.

*A. Experimental Setup*

**Implementation Details.** The hidden feature dimensions of encoder $D_{enc}$ and decoder $D_{dec}$ are set as 256 and 512, respectively. All the self-attention and cross-attention modules adopt a standard setting, i.e., six attention layers, each with eight heads. Dropout rate in attention layers is 0.1. Due to the sparsity of lane segments in CARLA, we deleted the dynamic map collection module in the original MTR.

Regarding endpoint intentions, sixteen motion queries ($N_{end}$ =16) were generated by clustering the endpoints on the training set of our proposed safety-critical dataset based on a k-means clustering algorithm. For risk intentions, we manually set a series of values to represent different urgency levels of traffic scenes, given the actual risk distribution. Specifically, we adopt two schemes: a model with three risk queries ($N_{risk}$ = 3), i.e., 300, 600, and 999 (maximum risk value representing collision conditions), and another model with four risk queries ($N_{risk}$ = 4), i.e., 300, 500, 650, and 999.

TABLE III
PERFORMANCE COMPARISON ON THE CLASSICAL TRAJECTORY PREDICTION METRICS.

| Method | Non-collision | | | | Collision-in-1s | | | | Collision-in-2s | | | | Collision-in-5s | | | |
|---|---|---|---|---|---|---|---|---|---|---|---|---|---|---|---|---|
| | ADE[a] | FDE | MR | b-FDE | ADE | FDE | MR | b-FDE | ADE | FDE | MR | b-FDE | ADE | FDE | MR | b-FDE |
| CV | 1.884 | 5.412 | 0.697 | - | 0.124 | 0.276 | 0.114 | - | 1.387 | 4.120 | 0.545 | - | 4.891 | 14.932 | 0.833 | - |
| CA | 3.373 | 9.535 | 0.809 | - | 0.218 | 0.580 | 0.431 | - | 2.520 | 7.247 | 0.773 | - | 7.659 | 22.307 | 0.905 | - |
| LSTM | 1.894 | 3.597 | 0.545 | 4.396 | 0.365 | 0.706 | 0.447 | 1.598 | 1.358 | 2.676 | 0.414 | 3.446 | 5.692 | 19.147 | 0.781 | 24.961 |
| GRU | 1.309 | 2.730 | 0.333 | 3.472 | 0.228 | 0.413 | 0.314 | 1.103 | 0.954 | 2.050 | 0.341 | 2.800 | 3.921 | 13.671 | 0.619 | 14.371 |
| Wayformer [4] | 0.956 | 2.176 | 0.227 | 2.792 | 0.163 | 0.298 | 0.196 | 0.954 | 0.582 | 1.559 | 0.227 | 2.033 | 2.445 | 6.390 | 0.452 | 7.048 |
| QCNet [5] | 0.923 | 1.952 | 0.165 | **2.611** | 0.163 | 0.249 | 0.098 | 0.921 | 0.659 | 1.539 | 0.182 | 2.194 | 2.274 | 6.114 | 0.405 | 6.784 |
| EDA [43] | 0.892 | 2.341 | 0.273 | 3.121 | 0.131 | 0.286 | 0.098 | 0.918 | 0.600 | 1.601 | 0.227 | 2.289 | 2.006 | 4.871 | 0.286 | 5.748 |
| HPNet [7] | 0.815 | 1.870 | **0.156** | 2.715 | 0.160 | 0.259 | 0.176 | 0.954 | 0.626 | 1.580 | 0.159 | 2.154 | 2.491 | 5.683 | 0.286 | 6.338 |
| MTR [10] | 0.888 | 2.167 | 0.234 | 2.856 | 0.123 | 0.223 | 0.098 | 0.958 | 0.593 | 1.471 | 0.204 | 2.223 | 1.940 | 5.186 | 0.357 | 5.820 |
| MTR-e2e [10] | 1.101 | 3.001 | 0.407 | 3.504 | 0.136 | 0.290 | 0.196 | **0.820** | 0.681 | 1.942 | 0.273 | 2.436 | 2.772 | 8.441 | 0.571 | 8.912 |
| Ours-3rq | **0.752** | 1.882 | 0.164 | 2.629 | **0.101** | **0.176** | **0.059** | 0.932 | **0.470** | **1.203** | 0.182 | **1.947** | **1.757** | **4.601** | **0.262** | **5.380** |
| Ours-4rq | 0.760 | **1.849** | 0.190 | 2.621 | 0.110 | 0.210 | 0.078 | 0.873 | 0.485 | 1.237 | **0.114** | 1.974 | 1.891 | 4.656 | 0.333 | 5.446 |

a. ADE (minADE), FDE (minADE), MR (Miss Rate), and b-FDE (brier-minFDE) all represent multimodal prediction performance with six trajectories (i.e., $k = 6$). The best entry for a metric is marked in bold.

**Training Details.** The proposed model was trained for 200 epochs using an AdamW optimizer with a weight decay of 0.01. The initial learning rate was set as $1 \times 10^{-4}$ and decayed by a factor of 0.5 at the epoch of 100, 125, 150, and 175. The model was trained with a batch size of 64 on eight NVIDIA GeForce RTX 3090 GPUs.

*B. Quantitative Results*

**Comparison with Baselines.** We compared the proposed risk-aware trajectory prediction model's performance with several baseline models, depicted as follows:

- **Physical-based Models**: constant velocity model (**CV**) and constant acceleration model (**CA**). They predict future trajectories based on the target vehicle's current kinematic state by assuming the agent moves at a constant velocity or acceleration.
- **Classic Machine Learning Models**: long short-term memory (**LSTM**) and gated recurrent unit (**GRU**).

**SOTA Trajectory Prediction Models**: Motion transformer (**MTR**) and its end-to-end variant, **MTR-e2e** [10], [37], **Wayformer** [4], Query-Centric trajectory prediction network (**QCNet**) [5], trajectory prediction network with Evolving and Distinct Anchors (**EDA**) [43], and trajectory prediction network with Historical Prediction attention (**HPNet**) [7]. For a fair comparison, we re-trained all the above SOTA models on the proposed safety-critical trajectory prediction dataset, just like ours. For each model, we compared the prediction performance of models with or without pre-trained weights imported, and the model with the best performance was selected.

**Performance on Classical Metrics.** Based on the proposed safety-critical trajectory prediction dataset, we first evaluated our models using classical metrics. As shown in Table III, the proposed two models, i.e., ours-3rq and ours-4rq, outperformed the baseline methods on most metrics with remarkable margins, averagely decreasing minADE and minFDE by 13.6% (i.e., 7.7%, 17.9%, 19.2%, and 9.4% across the four groups) and 11.5% (i.e., 1.1%, 21.1%, 18.2%, and 5.5% across the four groups) from the ten baseline models' best performance. From the total 16 metrics, our proposed models only showed suboptimal results on the metrics of MR and b-FDE for Non-collision and b-FDE for collision-in-1s, yet they still remained very close to optimal performance.

Traditional physical-based methods, i.e., CV and VA, performed the worst as expected, especially in the long-term prediction horizon, e.g., Non-collision and Collision-in-5s. However, CV has achieved pretty good performance for short-term prediction, e.g., Collision-in-1s, even better than some SOTA trajectory prediction models. This proves that it is feasible for some existing studies to use the CV model to predict vehicle trajectories in the event of an impending collision, where a collision generally occurs no more than 1s [26], [44]. As classic machine learning models, GRU performed better than LSTM, but both performances were far inferior to that of the recently proposed models due to their simple network structure. As for the SOTA models, the latest proposed methods (e.g., EDA and HPNet) did not exhibit a significant performance improvement over the older methods (e.g., MTR and Wayformer). This could be caused by the fact that the latest methods made elaborative modifications tailored to a specific benchmark to obtain performance improvement in the leaderboard, which thus might compromise their generalization ability to some extent. As a vital basis that derived many other SOTA models (e.g., EDA), MTR demonstrated good generalization ability with superior performance among the several SOTA models. Thus, our proposed models were developed based on MTR. Compared with these previous SOTA models, our risk-aware model achieved significantly better performance on most classical metrics by incorporating various architecture modifications and training techniques tailored to safety-critical scenarios.

TABLE IV
PERFORMANCE COMPARISON ON THE METRICS TAILORED TO SAFETY-CRITICAL SCENARIOS.

| Method | Traffic Risk | | | Collision Velocity | | | Collision Time | | | Collision Miss Rate | |
|---|---|---|---|---|---|---|---|---|---|---|---|
| | $\text{MSE}_{\text{risk}}^{k=1}$ | $\text{MSE}_{\text{risk}}^{k=6}$ | $\text{MR}_{\text{risk}}^{k=6}$ | $\text{MSE}_{\text{velo}}^{k=1}$ | $\text{MSE}_{\text{velo}}^{k=6}$ | $\text{MR}_{\text{velo}}^{k=6}$ | $\text{MSE}_{\text{time}}^{k=1}$ | $\text{MSE}_{\text{time}}^{k=6}$ | $\text{MR}_{\text{time}}^{k=6}$ | $\text{MR}_{\text{coll}}^{k=1}$ | $\text{MR}_{\text{coll}}^{k=6}$ |
| CV | 161.6 | 161.6 | 0.475 | 5.348 | 5.348 | 0.555 | 34.629 | 34.629 | 0.438 | 0.320 | 0.320 |
| CA | 205.7 | 205.7 | 0.557 | 6.358 | 6.358 | 0.642 | 46.423 | 46.423 | 0.531 | 0.443 | 0.443 |
| LSTM | 235.1 | 135.9 | 0.438 | 5.421 | 3.989 | 0.459 | 59.224 | 30.191 | 0.394 | 0.521 | 0.405 |
| GRU | 234.5 | 115.1 | 0.426 | 5.621 | 3.418 | 0.350 | 62.794 | 25.969 | 0.339 | 0.588 | 0.361 |
| Wayformer [4] | 123.4 | 60.3 | 0.242 | 4.018 | 2.768 | 0.263 | 26.479 | 17.665 | 0.222 | 0.252 | 0.175 |
| QCNet [5] | 120.7 | 46.2 | 0.181 | 4.539 | 2.299 | 0.285 | 28.809 | 11.649 | 0.170 | 0.273 | 0.093 |
| EDA [43] | 118.4 | 49.9 | 0.217 | 3.794 | 2.222 | 0.255 | 24.077 | 11.407 | 0.165 | 0.232 | 0.093 |
| HPNet [7] | 114.6 | 43.3 | 0.177 | 4.598 | 2.182 | 0.248 | 23.737 | 9.994 | 0.165 | 0.211 | 0.092 |
| MTR [10] | 104.3 | 48.0 | 0.182 | 3.883 | 2.333 | 0.248 | 25.160 | 12.856 | 0.191 | 0.237 | 0.103 |
| MTR-e2e [10] | 258.5 | 80.0 | 0.310 | 8.367 | 3.495 | 0.394 | 55.742 | 20.871 | 0.268 | 0.691 | 0.201 |
| Ours-3rq | 99.6 | 42.5 | 0.179 | 3.827 | **2.065** | **0.219** | 21.830 | 9.051 | 0.155 | **0.201** | **0.077** |
| Ours-4rq | **94.4** | **40.8** | **0.166** | **3.553** | 2.075 | 0.263 | 22.175 | 10.531 | **0.150** | 0.211 | 0.093 |

**Performance on Metrics Tailored to Safety-Critical Scenarios.** Table IV displays the performance comparison on metrics tailored to safety-critical scenarios. Our models consistently maintained the leading position on all the eleven metrics. This demonstrates that our models are not just good at predicting trajectories but also do well in estimating the future evolution of traffic risks and the consequences of collisions, e.g., collision velocity and collision time, which is essential for the subsequent guidance of autonomous vehicles to carry out correct safety decisions to reduce collision probability and mitigate injuries [26], [45], [46].

In addition, note that MTR outperformed HPNet on most classical metrics, but HPNet demonstrated much better performance on most of our proposed safety-critical metrics. This displays a potential risk in selecting the best trajectory prediction model for safety-critical scenarios solely based on the classical trajectory prediction metrics, further proving the importance of safety-critical metrics. Moreover, we found MTR performed relatively well for unimodal prediction, i.e., $k=1$, but showed an apparent performance reduction for multimodal prediction with $k=6$, which could be attributed to the inappropriate design of multimodal queries. To rectify this defect, according to the characteristics of safety-critical scenarios, we innovatively incorporated a set of risk intention queries and combined them with endpoint intentions to guide more risk-aware multimodal predictions.

**Performance Comparison Between the Two Proposed Models.** The performance of the two models was similar on most evaluation metrics. One difference is that the model of ours-3rq performed better on most classical metrics, while ours-4rq demonstrated slightly better predictions on safety-critical metrics, which can be attributed to the refined design of risk intention queries, i.e., incorporating more risk levels.

**Performance on the Estimation of Collision Probability.** Some existing studies on traffic safety utilized Monte Carlo to randomly sample agents' future trajectories and then estimated their collision probability under safety-critical scenarios [47], [48]. Inspired by these approaches, we estimated the collision probability in a similar way and substituted the randomly sampled trajectories with the predicted multimodal trajectories of the target vehicle. The overall estimated collision probability was 54.7% for the model ours-4rq. Further, the collision probability was estimated as 78.8% and 28.0% for collision and non-collision cases, respectively. For ours-3rq, the three probabilities were 55.0%, 78.9%, and 28.3%. The significant probability difference between collision cases and non-collision cases proved that our proposed models could be used for autonomous vehicles to detect whether there is a potential collision in the near future.

### C. Qualitative Results

Besides quantitative results, we also provide qualitative results to show our predictions under safety-critical scenarios.

**Trajectory Prediction Visualization.** As shown in Fig. 7, four representative safety-critical scenarios are displayed, including two collision scenarios (I and II) and two non-collision scenarios (III and IV).

In Scenario I (Fig. 7a), the blue vehicle suddenly braked to stop when passing through a green light intersection, triggering traffic danger. Due to the small car-following distance and the driver's slow reaction, the target vehicle rear-ended the front vehicle. Seeing from the predicted trajectories represented by green dashed lines, our model accurately captured the target vehicle's intention of longitudinal braking. Additionally, our model also predicted two reasonable turning intention modes, demonstrating the comprehensiveness of the model in multimodal prediction. The risk evolution curves below show that these turning intentions effectively reduced traffic risks, ultimately achieving collision avoidance.

Scenario II and III each presented a safety-critical scenario triggered by surrounding vehicles' malicious lane changes and deceleration (Fig. 7b and 7c). In Scenario II, the excessive relative speed between two vehicles led to an accident, while the target vehicle in Scenario III took deceleration in advance and avoided crashes. The predicted trajectories in Scenario II could be classified into two collision-avoidance intentions, namely, changing lanes to the right (with four trajectories, three of which still resulted in accidents) and driving out of the road to the left (with two trajectories, both of which avoided collisions). In Scenario III, our model accurately predicted the target vehicle' trajectory to evade collisions by timely braking. Additionally, our model also precisely estimated the speeds of the vehicles in the left front and left rear when predicting the two trajectories of turning left, thereby these trajectories adopting appropriate steering rates and deceleration without causing new accidents.

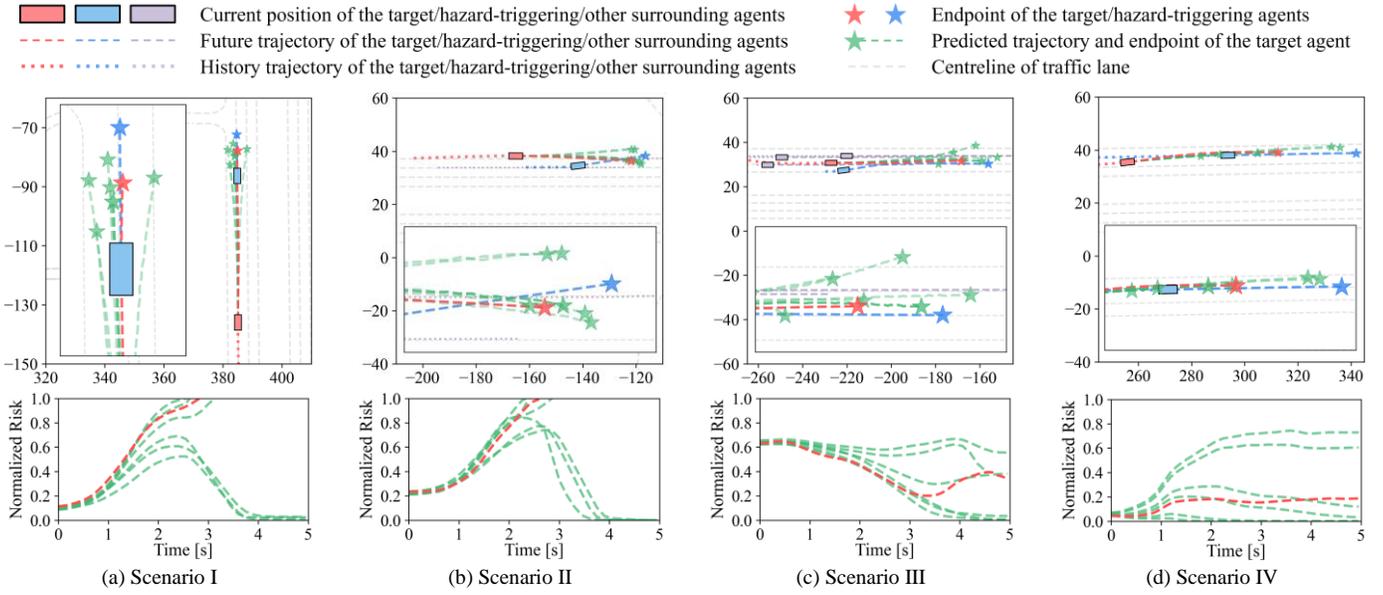

**Fig. 7.** Examples of multimodal trajectories predicted by our model under safety-critical scenarios. The first row displays the visualization of predicted trajectories, while the second row shows the predicted traffic risk (after normalization). In the first row, the target agent (red), hazard-triggering agent (blue), and other surrounding agents (purple) are color-coded for clarity. Future trajectory and historical trajectory are represented by dotted lines and dashed lines, respectively. Predicted multimodal trajectories are in green. Stars denote trajectory endpoints. Gray dashed lines represent centrelines of traffic lanes. For brevity, surrounding vehicles that are less relevant to the predicted results are hidden. The second row displays the ground-truth and predicted traffic risks in red and green, respectively. The normalized risk value of 1 represents the occurrence of collisions.

In Scenario IV (Fig. 7d), the target vehicle was instructed to change lanes to the left, but simultaneously, the blue vehicle began emergency braking to create a hazard. In the predicted trajectories, the target vehicle adopted different longitudinal decelerations to reduce collision risks to varying degrees, while two other predicted trajectories also avoided collisions without significant deceleration, which were achieved by executing another left lane change.

Overall, we concluded that our proposed model achieved ideal multimodal prediction capabilities under different safety-critical scenarios and various risk levels. It not only achieved a high degree of overlap with the ground-truth trajectories but also accurately captured the future evolution of traffic risks.

**Qualitative Comparison with Baselines.** We also visualized the comparison with SOTA models (Fig. 8).

In Scenario I (Fig. 8a), trajectories predicted by different models all collided with the blue vehicle, but only our model and EDA predicted the target vehicle's deceleration, achieving a more accurate estimation of the collision speed. In Scenario

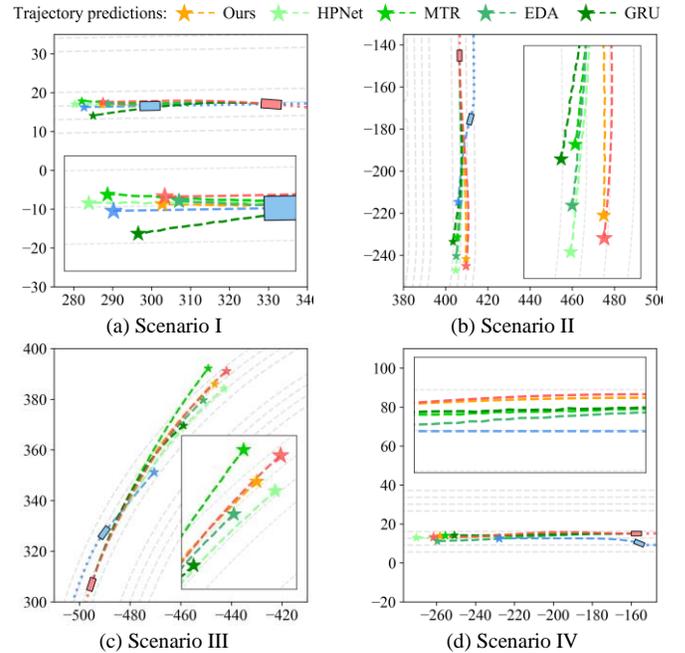

**Fig. 8.** Qualitative comparisons of our proposed model and other methods. For each model, only the predicted trajectory with the highest prediction probability is shown for brevity.

II (Fig. 8b), our model successfully predicted that the target vehicle would change lanes to the left while emergency braking to avoid the accident, while the other models only predicted the deceleration behaviors, still causing accidents. In the curved road shown in Scenario III (Fig. 8c), our model accurately captured the target vehicle's lane change intention.

TABLE V
EFFECTS OF THE CORE COMPONENTS OF THE PROPOSED RISK-AWARE TRAJECTORY PREDICTION FRAMEWORK.

| Core Components | | | Classical Metrics | | | | | | | | Tailored Metrics | | | | |
|---|---|---|---|---|---|---|---|---|---|---|---|---|---|---|---|
| RSE | ERQ | APR | Non-collision | | Collision-in-1s | | Collision-in-2s | | Collision-in-5s | | Traffic Risk | | Collision Velocity | | Miss Rate |
| | | | ADE | FDE | ADE | FDE | ADE | FDE | ADE | FDE | $MSE_{risk}^{k=6}$ | $MR_{risk}^{k=6}$ | $MSE_{velo}^{k=6}$ | $MR_{velo}^{k=6}$ | $MR_{coll}^{k=6}$ |
| | | | 0.888 | 2.167 | 0.123 | 0.223 | 0.593 | 1.471 | 1.940 | 5.186 | 48.0 | 0.182 | 2.333 | 0.248 | 0.103 |
| ✓ | | | 0.843 | 2.033 | 0.121 | 0.250 | 0.542 | 1.523 | 1.990 | 5.012 | 46.9 | 0.193 | 2.218 | 0.241 | 0.113 |
| ✓ | ✓ | | 0.784 | 1.979 | 0.133 | 0.268 | 0.540 | 1.439 | **1.842** | 5.107 | 44.4 | 0.174 | 2.271 | **0.234** | 0.098 |
| ✓ | ✓ | ✓ | **0.760** | **1.849** | **0.110** | **0.210** | **0.485** | **1.237** | 1.891 | **4.656** | **40.8** | **0.166** | **2.075** | 0.263 | **0.093** |

In Scenario IV (Fig. 8d), the trajectory endpoints predicted by different models were all very close, but except for our model, the other models all experienced collision accidents along the way. After zooming in on the predicted trajectories, we found that only our model captured a slight rightward turn in the target vehicle's ground-truth trajectory, successfully avoiding a scrape collision with the blue vehicle.

### D. Ablation Studies

**Effectiveness of Risk-Aware Modules.** Following the same routine as the main experiment, we conducted a series of ablation experiments for the three proposed risk-aware components, i.e., RSE, ERQ, and APR. Considering the sequence of the three components in processing input data, we added them to the original MTR model to observe their impact on prediction accuracy. As displayed in Table V, most evaluation metrics, e.g., non-collision minADE and $MSE_{risk}^{k=6}$, kept reducing when different components were added. The introduction of RSE improved the encoder's scene encoding ability to understand safety-critical scene context, leading to the reduction of some classical metrics, e.g., minADE and minFDE for non-collision cases. As the core module of our model, ERQ was designed to achieve risk-aware predictions by incorporating a set of pre-defined risk intention queries as prediction priors, eventually contributing to the significant reduction on most metrics, especially for collision-in-5s minADE and $MR_{velo}^{k=6}$. Introducing APR further improved the whole technical framework with the best performance. Overall, the ablation studies preliminarily proved that each component contributes to the final enhancement of trajectory prediction performance under safety-critical scenarios.

**Effects of the Number of Risk Intention Queries.** Two models with three and four risk queries were proposed in the main experiment, i.e., ours-3rq and ours-4rq. Table VI further investigated the effects of the number of risk intention queries (from two to six). Models of ours-3rq and ours-4rq performed best in most metrics. When only two risk queries exist, the model performed well on classical trajectory-related metrics and even obtained the lowest minFDE for the collision-in-5s cases. However, its performance on the tailored safety-critical metrics declined significantly due to insufficient consideration of the risk levels. The prediction performance also dropped when the number of risk queries improved to five or six. This might be caused by the lack of training data when the modes of multimodal prediction (i.e., $N_{risk} \times N_{end}$) improve with the number of risk intention queries.

TABLE VI
EFFECTS OF THE NUMBER OF RISK INTENTION QUERIES.

| Query Number | Non-collision | | Collision-in-5s | | Tailored Metrics | | |
|---|---|---|---|---|---|---|---|
| | ADE | FDE | ADE | FDE | $MR_{risk}^{k=6}$ | $MR_{velo}^{k=6}$ | $MR_{coll}^{k=6}$ |
| 2 | 0.782 | 1.935 | 1.862 | **4.393** | 0.214 | 0.255 | 0.134 |
| 3 | **0.752** | 1.882 | **1.757** | 4.601 | 0.179 | **0.219** | **0.077** |
| 4 | 0.760 | **1.849** | 1.891 | 4.656 | **0.166** | 0.263 | 0.093 |
| 5 | 0.753 | 1.866 | 1.884 | 5.196 | 0.185 | 0.241 | 0.108 |
| 6 | 0.781 | 1.922 | 1.909 | 4.941 | 0.196 | 0.241 | 0.113 |

TABLE VII
EFFECTS OF THE WEIGHT OF THE RISK PREDICTION LOSS.

| Risk Weight | Non-collision | | Collision-in-5s | | Tailored Metrics | | |
|---|---|---|---|---|---|---|---|
| | ADE | FDE | ADE | FDE | $MR_{risk}^{k=6}$ | $MR_{velo}^{k=6}$ | $MR_{coll}^{k=6}$ |
| 1 | 0.798 | 1.920 | 1.799 | 4.697 | **0.166** | 0.241 | 0.093 |
| 0.3 | **0.752** | 1.882 | **1.757** | **4.601** | 0.179 | **0.219** | **0.077** |
| 0.1 | 0.776 | 1.889 | 1.780 | 4.658 | 0.182 | 0.248 | 0.098 |
| 0.03 | 0.760 | **1.851** | 1.762 | 4.612 | 0.217 | 0.263 | 0.113 |

**Effects of the Weight of Risk Prediction Loss $\mathcal{L}_{risk}$.** Table VII investigated the effects of the relative weight of risk prediction loss. The weight of 0.3 performed best on most metrics. Additionally, we found that increasing risk weight was helpful in improving the forecasting effect on the metrics tailored to safety-critical scenarios (e.g., $MR_{risk}^{k=6}$ of 0.166 with the risk weight of 1), and vice versa was conducive to the performance on classical metrics (e.g., non-collision minFDE of 1.851 with the risk weight of 0.03).

### E. Additional Experiments on Risk Queries

To further analyze the effect of the proposed risk intention queries on safety-critical trajectory prediction, we made an additional statistical analysis of the distribution characteristics of the multimodal trajectory predictions guided by various risk queries. The results are shown in Table VIII. We observed that for both models, i.e., ours-3rq and ours-4rq, the estimated crash rate and maximum traffic risk monotonically rose with the increase of the risk query level. We considered that through carefully designing a set of pre-defined risk values as prediction priors (from low-risk level to high-risk level), after training, different risk intention queries successfully learned to be responsible for their respective risk regions when predicting multimodal trajectories, eventually contributing to the achievement of risk-aware trajectory prediction under safety-critical scenarios.

TABLE VIII
CHARACTERISTICS OF THE MULTIMODAL TRAJECTORY PREDICTIONS GUIDED BY VARIOUS RISK QUERIES

| Risk Query Level | Crash Rate | | Maximum Traffic Risk | |
|---|---|---|---|---|
| | Ours-3rq | Ours-4rq | Ours-3rq | Ours-4rq |
| I | 39.3% | 28.9% | 528.4 | 456.2 |
| II | 47.3% | 41.9% | 590.5 | 513.6 |
| III | 52.4% | 51.3% | 615.3 | 618.3 |
| IV | - | 53.5% | - | 640.9 |

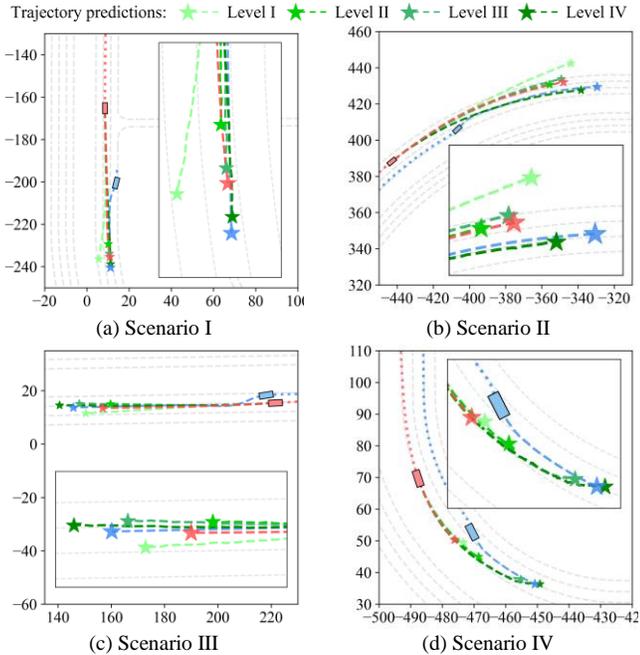

**Fig. 9.** Multimodal trajectory predictions under the guidance of risk intention queries with various risk levels.

We further visualized four specific scenarios in Fig. 9 to analyze the effects of various risk queries. On the whole, trajectory predictions under the guidance of higher-risk-level queries tended to be more aggressive and more likely to cause collision accidents, e.g., the trajectory shown by the dark green line in Fig. 9b, which refused to change lanes, remained in its original lane continuously, and ended up in a rear-end collision. On the contrary, trajectories with lower-risk-level queries were more conservative with the first goal of staying away from the dangerous objects as much as possible, e.g., the light-green trajectory in Fig. 9a, which veered out of the lane in advance to avoid the hazard-triggering vehicles. We found an interesting phenomenon: when faced with the safety-critical scenario shown in Fig. 9d, the predicted trajectories guided by high-risk-level queries even opted for a slight acceleration, seemingly to trigger a collision deliberately.

## VII. CONCLUSION

This study presents a novel risk-aware vehicle trajectory prediction framework tailored to safety-critical scenarios. Different from conventional approaches, we propose a series of risk-aware prediction components specifically developed in combination with the characteristics of safety-critical scenarios. The proposed risk-incorporated scene encoder aims to achieve risk-aware encoding of hazardous scene contexts by introducing quantitative traffic risk computed based on an artificial potential field approach. Combining spatial endpoint intentions and traffic risk intentions into a two-dimensional query matrix, the multimodal trajectories predicted by our trajectory decoder effectively cover various motion modes and risk levels. The effectiveness of our model is verified through quantitative and qualitative analyses on a newly acquired safety-critical trajectory prediction dataset. The experimental results reveal that our model outperforms mainstream methods and existing SOTA models on not only the regular evaluation metrics (e.g., minFDE) but also the tailored metrics developed for safety-critical scenarios.

To the best of the authors' knowledge, this study is the first to conduct trajectory prediction under safety-critical scenarios. To sum up, the contributions of this study are threefold: (1) A series of core trajectory prediction components to adapt to safety-critical scenarios and achieve risk-aware predictions. (2) A new trajectory prediction dataset that provides large-scale driving behaviors under safety-critical scenarios. (3) A series of evaluation metrics tailored to trajectory prediction under safety-critical scenarios.

This study just focuses on trajectory prediction tasks under safety-critical scenarios. However, in real-world traffic scenarios, normal situations gradually evolve into hazardous ones. Thus, it is challenging to define the exact time when traffic hazards occur precisely. Thus, more efforts should be made to develop a trajectory prediction approach that can handle both normal and safety-critical scenarios by adaptively switching prediction modes according to the risk level in the scenario. To achieve it, domain adaptation may be a promising method. Meanwhile, compared with existing public datasets, our proposed safety-critical dataset is not big enough to bring the ability of data-driven methods into full play. The further expansion and detailed validation of our dataset are now underway.

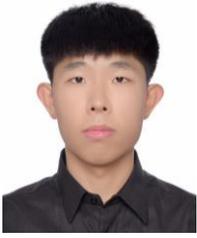

**Qingfan Wang** received the B.S. degree in mechanical engineering from the Harbin Institute of Technology, Harbin, China, in 2020. He is pursuing a Ph.D. degree with the State Key Laboratory of Intelligent Green Vehicle and Mobility, School of Vehicle and Mobility, Tsinghua University. He is currently a Visiting Scholar with the School of Mechanical and Aerospace Engineering, Nanyang Technological University. His research interests include autonomous driving, trajectory prediction, and intelligent occupant protection.

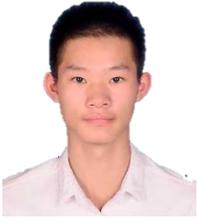

**Dongyang Xu** received the B.S. degree in mechanical engineering from the Xidian university, Xian, China, in 2022. He is currently pursuing a master's degree with the State Key Laboratory of Intelligent Green Vehicle and Mobility, School of Vehicle and Mobility, Tsinghua University. His research interests include end-to-end autonomous driving and motion planning.

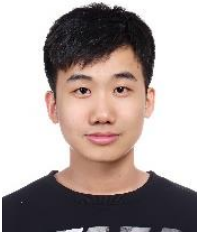

**Gaoyuan Kuang** received the B.S. degree in mechanical engineering from the Harbin Institute of Technology, Harbin, China, in 2022. He is currently pursuing a master's degree with the State Key Laboratory of Intelligent Green Vehicle and Mobility, School of Vehicle and Mobility, Tsinghua University. His research interests include virtual simulation, autonomous driving, and accelerated testing.

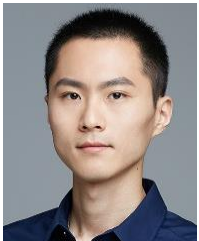

**Chen Lv** (Senior Member, IEEE) is an Associate Professor at School of Mechanical and Aerospace Engineering, and the Cluster Director in Future Mobility Solutions, Nanyang Technological University, Singapore. He joined NTU and founded the Automated Driving and Human-Machine System (AutoMan) Research Lab since June 2018. His research focuses on intelligent vehicles, automated driving, and human-machine systems, where he has published 4 books, over 100 papers, and obtained 12 granted patents. He serves as Associate Editor for IEEE T-ITS, IEEE TVT, and IEEE T-IV. He received many awards and honors, selectively including the IEEE IV Best Workshop/Special Session Paper Award in 2018, Automotive Innovation Best Paper Award in 2020, the winner of Waymo Open Dataset Challenges at CVPR 2021 and 2022, Machines Young Investigator Award, Nanyang Research Award (Young Investigator), SAE Ralph R. Teetor Educational Award, etc.

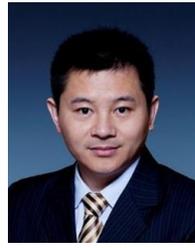

**Shengbo Eben Li** (Senior Member, IEEE) received the M.S. and Ph.D. degrees from Tsinghua University in 2006 and 2009. He worked at Stanford University, University of Michigan, and University of California, Berkeley. He is currently a tenured professor at Tsinghua University. His active research interests include intelligent vehicles and driver assistance, reinforcement learning and distributed control, optimal control and estimation, etc. He was the recipient of Best Paper Award in 2014 IEEE ITS Symposium, National Award for Technological Invention in China (2013), Excellent Young Scholar of NSF China (2016), Young Professorship of Changjiang Scholar Program (2016). He is now the IEEE senior member and serves as associated editor of IEEE ITSM and IEEE Trans. ITS, etc.

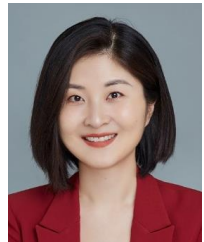

**Bingbing Nie** (Member, IEEE) received the B.S. degree from Tsinghua University, China, in 2007, the M.S. degree from RWTH-Aachen, Germany, in 2009, and the Ph.D. degree from Tsinghua University in 2013. She is an Associate Professor at the School of Vehicle and Mobility, Tsinghua University, China. Prior to joining Tsinghua University, she worked at General Motors R&D and University of Virginia. She has co-authored more than 70 technical papers and is a co-inventor of over 20 patents. Her active research areas include vehicle safety, applied biomechanics, and intelligent protection. She has served as IRCOBI Council Member, Session Organizer of Pedestrian and Cyclist Safety of SAE World Congress, and Secretary of SAE Vehicle Safety Technical Committee.